\documentclass{article}

\PassOptionsToPackage{numbers, compress}{natbib}

\usepackage[preprint]{neurips_2019}

\usepackage[utf8]{inputenc} 
\usepackage[T1]{fontenc}    
\usepackage{hyperref}       
\usepackage{url}            
\usepackage{booktabs}       
\usepackage{amsfonts}       
\usepackage{nicefrac}       
\usepackage{microtype}      

\usepackage{multirow}
\usepackage{color}
\usepackage{amsmath,amssymb}
\usepackage{graphicx}
\usepackage{wrapfig}
\usepackage{lipsum}
\usepackage{natbib}
\usepackage{caption}
\usepackage{subcaption}

\DeclareMathOperator*{\argmax}{arg\,max}

\newif\ifcomments
\commentsfalse
\commentstrue  
\ifcomments
\newcommand{\asma}[1]{\textcolor{cyan}{\bf\small [#1 --AG]}}
\newcommand{\judy}[1]{\textcolor{blue}{\bf\small [#1 --JS]}}
\newcommand{\natasha}[1]{\textcolor{magenta}{\bf\small [#1 --Natasha]}}
\newcommand{\cut}[1]{\textcolor{red}{\bf\small [#1 --cut?]}}
\newcommand{\agata}[1]{\textcolor{green}{\bf\small [#1 --Agata]}}
\else
\newcommand{\asma}[1]{}
\newcommand{\judy}[1]{}
\newcommand{\natasha}[1]{}
\newcommand{\cut}[1]{}
\newcommand{\agata}[1]{}
\fi

\title{Way Off-Policy Batch Deep Reinforcement Learning \\
      of Implicit Human Preferences in Dialog}

\author{ 
  Natasha Jaques, Asma Ghandeharioun, Judy Hanwen Shen, Craig Ferguson, \\ 
  \textbf{Agata Lapedriza, Noah Jones, Shixiang Gu, Rosalind Picard} \\
  Department of Media Arts and Science\\
  Massachusetts Institute of Technology\\
  Cambridge, MA 02139 \\
  \texttt{jaquesn@mit.edu} \\
}

\begin{document}

\maketitle

\begin{abstract}

Most deep reinforcement learning (RL) systems are not able to learn effectively from off-policy data, especially if they cannot explore online in the environment. These are critical shortcomings for applying RL to real-world problems where collecting data is expensive, and models must be tested offline before being deployed to interact with the environment -- e.g. systems that learn from human interaction. Thus, we develop a novel class of off-policy batch RL algorithms, which are able to effectively learn offline, without exploring, from a fixed batch of human interaction data. We leverage models pre-trained on data as a strong prior, and use KL-control to penalize divergence from this prior during RL training. We also use dropout-based uncertainty estimates to lower bound the target Q-values as a more efficient alternative to Double Q-Learning. The algorithms are tested on the problem of open-domain dialog generation -- a challenging reinforcement learning problem with a 20,000-dimensional action space. Using our Way Off-Policy algorithm, we can extract multiple different reward functions post-hoc from collected human interaction data, and learn effectively from all of these. We test the real-world generalization of these systems by deploying them live to converse with humans in an open-domain setting, and demonstrate that our algorithm achieves significant improvements over prior methods in off-policy batch RL.

\end{abstract}

\section{Introduction}
In order to scale deep reinforcement learning (RL) to safety-critical, real-world domains, two abilities are needed. First, since collecting real-world interaction data can be expensive and time-consuming, algorithms must be able to leverage off-policy data -- collected from vastly different systems, far into the past -- in order to learn. Second, it is often necessary to carefully test a policy before deploying it to the real world; for example, to ensure its behavior is safe and appropriate for humans. Thus, the algorithm must be able to learn offline first, from a static batch of data, without the ability to explore. 

This off-policy, \textit{batch reinforcement learning} (BRL) setting represents a challenging RL problem. Most deep RL algorithms fail to learn from data that is not heavily correlated with the current policy \cite{fujimoto2018off}. Even models based on off-policy algorithms like $Q$-learning fail to learn when the model is not able to explore during training. This is due to the fact that such algorithms are inherently optimistic in the face of uncertainty. When value estimates are noisy, taking the maximum over estimates of future reward leads to a persistent overestimation bias. In a normal RL setting, this drives the model to explore areas of the state-action space for which the value estimates have the highest variance, thus enabling it to refine them. In a batch setting where the model cannot explore, it is instead driven to value parts of the state-action space for which it has little to no data to learn a good policy.

We propose to resolve these issues by leveraging a pre-trained generative model of the state-action space trained on known sequences of interaction data. While training with RL, we penalize divergence from this prior model with different forms of KL-control. 
We benchmark against a discrete adaptation of Batch Constrained Q (BCQ) \cite{fujimoto2018off}, a recently proposed BRL algorithm for continuous domains, and show that KL-control achieves superior performance.  Finally, we propose using dropout to obtain uncertainty estimates of the target Q values, and use this lower bound to alleviate the Q-learning overestimation bias. This provides a more efficient alternative to Clipped Double Q-Learning \cite{fujimoto2018addressing}.  

We apply these algorithms to a challenging, under-explored, real-world reinforcement learning problem: using implicitly expressed human reactions in chat to improve open-domain dialog systems. When a machine learning system interacts with humans, ideally we would like to learn about the humans' preferences in order to improve the performance of the system. Yet having humans manually indicate their preferences through explicit means like pressing a button (e.g. \cite{christiano2017deep}) or submitting a feedback report, does not scale. Instead, we would like to be able to use humans' implicit reactions, such as the sentiment they express, or the length of the conversation, in order to improve the policy. 

Applying off-policy batch RL to language generation is challenging because the number of potential combinations of words and sentences leads to a combinatorial explosion in the size of the state space. The action space -- the set of frequent vocabulary words in the English language -- is 20,000-dimensional. This compounds the overestimation problem, making BRL even more difficult. However, when learning from human interactions in the wild, it is crucial to be able to learn offline and test the policy before deploying it, lest it learn inappropriate behaviors (e.g. \cite{tay}).

To support this work, we developed an interactive online platform that allows humans to chat with deep neural network dialog models running on GPU; the BRL models trained for this study are available live at \url{https://neural.chat/rl}. 
Through this platform we collected human responses to a set of over 40 different dialog models over the course of several months. Using our Way Off-Policy algorithm, we are able to effectively learn from this batch of data, in spite of the fact that it was generated with a vastly different set of model architectures, which were trained on different  datasets. Further, we use the batch to learn from many different reward functions designed post-hoc to extract implicit human preferences, something that is only possible with effective off-policy BRL. 


\section{Related Work}
The approach we propose is based on KL-control, a branch of stochastic optimal control (SOC)~\citep{stengel1986stochastic} where the Kullback-Leibler (KL) divergence from some distribution is used to regularize an RL policy (e.g. \citep{abdolmaleki2018maximum, kappen2012optimal,rawlik2012stochastic, todorov2006linearly}). Well-known examples include Trust Region Policy Optimization (TRPO) \cite{slmja-trpo-15}, and use conservative, KL-regularized policy updates to restrict the RL algorithm to stay close to its own prior policy (e.g. ~\citep{haarnoja2018soft,kakade2002natural,peters2010relative,rawlik2012stochastic}). 
KL-control can also be applied to entropy maximization (e.g. \cite{ziebart2008maximum}); for example, $G$-learning penalizes KL-divergence from a simple uniform distribution in order to cope with overestimation of $Q$-values \cite{fox2016taming}. Soft $Q$-learning motivates using a Boltzmann distribution in the value function as a way of performing maximum entropy RL \cite{haarnoja2017reinforcement}. 
KL-control has also been used to improve transfer learning between maximum likelihood estimation (MLE) training on data, and training with RL \cite{jaques2017sequence}. To the best of our knowledge, our work is the first to propose KL-control as a way of improving off-policy learning without exploration in a BRL setting.

Other strategies to improve off-policy learning have been proposed, although many focus on scenarios where the policy is able to explore and collect more data (e.g. \cite{degris2012off,riedmiller2005neural}). 
In the deep RL setting, policy gradients can be corrected to account for the difference in the distribution of states visited under the original policy and the learned off-policy algorithm \cite{liu2019off}. Covariance-shift-based methods have been adapted to the off-policy deep RL setting to deal with the issue of value divergence \cite{gelada2019off}. Normalized feature representations have been proposed as an alternative approach \cite{bhatt2019crossnorm}. 
Batch Constrained Q-learning (BCQ) \cite{fujimoto2018off} tackles off-policy batch learning in continuous action domains by training a generative model of the batch, $p(a|s)$, sampling from this model, and selecting the best action based on a $Q$-estimate. This approach fails to integrate information about the distribution $p(a|s)$ directly into the policy, and cannot scale to scenarios in which the state-action space is large, and the amount of available batch data is too small to train $p(a|s)$. Many works from off-policy policy evaluation use importance sampling or model estimation to investigate the problem of estimating the performance of a policy given a batch of off-policy data (e.g. \cite{farajtabar2018more,jiang2016doubly,precup2000eligibility,thomas2016data}). Effective off-policy learning gives us the ability to learn from many different rewards post-hoc, something that could potentially improve techniques which use the relabeling trick (e.g. \cite{kaelbling1993learning,andrychowicz2017hindsight}).

We propose using dropout to approximate model uncertainty of the target $Q$-network. The idea of using dropout to estimate uncertainty in neural networks was first proposed by Gal and colleagues (2016) \cite{gal2016dropout}. Different forms of uncertainty estimates have been used in RL (e.g. \cite{kahn2017uncertainty, osband2016deep}); for example, Bayesian uncertainty estimates have been proposed as an alternative to double DQN \cite{azizzadenesheli2018efficient}. 

Improving dialog systems with RL has largely been restricted to task-oriented dialog systems, which have a limited number of task-specific actions (e.g. \cite{fatemi2016policy, gavsic2011line, liu2017iterative, liu2018dialogue, su2017sample}). 
These approaches may incorporate human input, usually through explicit, manual feedback (e.g. \cite{shah2018bootstrapping}), but sometimes with more implicit signals, such as the user interrupting the system or starting over \cite{shi2018sentiment}. Attempts to expand RL to the open-domain dialog setting are less numerous. Even in this setting, authors may choose to use a highly restricted action space; for example, using RL to choose which scripted or MLE dialog model to invoke to answer a user's query \cite{serban2017deep}. Early attempts to apply deep RL to the full vocabulary-sized action space relied mainly on hand-crafted rewards that described qualities of the generated text, such as \textit{ease of answering} \cite{li2016deep}. This approach has been extended to use a discriminator trained to distinguish human from generated text as a reward function 
\cite{li2017adversarial,li2018dialogue}. While some work has incorporated implicit signals such as sentiment \cite{hancock2019learning} and conversation length \cite{zhou2018design} in MLE systems, the idea of using such signals as a reward for RL is relatively unexplored. Shin and colleagues uses on-policy learning in conjunction with a user-sentiment approximator to improve a seq2seq model \cite{shin2019happybot}, but are unable to learn directly from user feedback. To the best of our knowledge, we are the first to use batch RL to train hierarchical open-domain dialog models on implicit cues gained from real human interactions.

\section{Methods}
\label{sec:methods}
We employ typical RL notation in which $s_t$ represents the environment state at time $t$, the agent takes action $a_t$ according to its policy $\pi(a_t|s_t)$, and receives a reward $r(s_t,a_t)$. 
The agent's goal is to maximize reward over an episode trajectory $\tau$, with a discount factor of $\gamma$ applied to future rewards. 
$Q$-learning methods learn an action-value estimate of the total expected discounted future reward, $Q_{\pi}(a_t,s_t) = \mathbb{E}_{\pi}[\sum_{t'=t}^{T} \gamma^{t'-t}r(s_{t'},a_{t'})]$, through iterative updates based on the Bellman equation:
\begin{align} 
\label{eq:bellman}
Q_{\theta_\pi}(s_t,a_t) &= r(s_t,a_t) + \gamma \mathbb{E}_{s_{t+1}\sim p(\cdot|s_t,a_t)}[\max_{a_{t+1}} Q_{\theta_T}(s_{t+1},a_{t+1})] 
\end{align}
In deep $Q$-learning \citep{dqn}, a $Q$-network approximates $Q_{\theta_\pi}(s_t,a_t)$ and drives the policy $\pi$. A second target $Q$-network approximates the expected reward from the next state, $Q_{\theta_T}(s_{t+1},a_{t+1})$ -- a standard practice for alleviating overestimation bias \cite{van2016deep}. 

To perform batch $Q$-learning, we first pre-train a generative model of $p(a|s)$ using a set of known environment trajectories. In our case, this model is then used to generate the batch data via human interaction. The weights of the $Q$-network and target $Q$-network are initialized from the pre-trained model, which helps reduce variance in the $Q$-estimates and works to combat overestimation bias. To train $Q_{\theta_{\pi}}$ we sample $<s_t, a_t, r_t, s_{t+1}>$ tuples from the batch, and update the weights of the $Q$-network to approximate Eq. \ref{eq:bellman}. This forms our baseline model, which we call \textit{Batch Q}.

\subsection{Dropout for uncertainty estimation of Target $Q$-values}
\label{sec:dropout}
Overestimation of $Q$-values becomes particularly problematic in the batch setting. The $Q$ estimates for state-action pairs which are not well covered in the batch will be noisy, and this variance will lead the \textit{max} operator in Eq. \ref{eq:bellman} to overestimate the value of these states. This drives the model to value regions of the state-action space for which it has no data to learn a reasonable policy, and no ability to explore to refine its estimates. Clipped Double $Q$-learning \cite{fujimoto2018addressing} addresses the overestimation problem by maintaining two independent pairs of $Q$-networks, 
and taking the minimum of their estimates of future reward. 
This approach is computationally expensive and memory intensive. Further, if following a transfer learning approach where the $Q$-network is initialized from a pre-trained MLE model (as we do in this paper), it is not clear how to obtain multiple independent target $Q$-networks.

Instead, we obtain a distribution over predictions from a single target $Q$-network trained with dropout, and take the lower bound of these to reduce overestimation bias. It has been shown that dropout approximates Bayesian uncertainty for neural networks, by assuming the weights of the network are drawn from a Gaussian prior,  $W~\sim~N(0,I)$, and using variational inference to estimate the posterior distribution $p(W|X, Y)$ \cite{gal2016dropout}. We perform dropout during both training and inference before each weight layer, and approximate the posterior such that the dropout distribution $q^W$ is a mixture of Gaussians, and $D_{KL}[q^W ||p(W|X, Y)]$ is minimized.
Given the target $Q$-network $Q_{\theta_T}$, we compute $Q(a_{t+1}, s_{t+1})$ using a Monte Carlo (MC) estimate of the lower-bound of $Q_{\theta_T}(a_{t+1}, s_{t+1})$ by running $M$ stochastic forward passes of the network, each with a new dropout mask $d_i \sim q^W$:
\begin{align}
    \label{eq:dropout_mc}
    Q(a_{t+1}, s_{t+1}) = \min_{i=1...M}[Q_{\theta_T}(a_{t+1}, s_{t+1}; d_i)]
\end{align}
Using the minimum operator penalizes high variance estimates, essentially leading the algorithm to be pessimistic in the face of uncertainty, rather than optimistic. Such a bias will push the model to favour actions that lead to states well covered by the batch data \cite{fujimoto2018off}. We evaluate the performance of this approach using a second baseline model, \textit{Batch Q MC}.

\subsection{Discrete Batch Constrained $Q$}
Batch Constrained Q-learning (BCQ) \cite{fujimoto2018off} proposes to address the BRL problem by constraining the actions of the $Q$-network to be close to the data contained within the batch. This is accomplished by learning a generative model of the batch, $G_w = p(a|s)$, and sampling from this model during learning and inference. Because BCQ is designed for continuous action domains, it applies a learned perturbation model $\xi(s,a;\Phi)$ which is allowed to alter the action within the range $[-\Phi, \Phi]$. BCQ learns $Q$-estimates that incorporate the perturbation model, $Q_{\theta}(s, a+\xi(s,a;\Phi))$. To act, $n$ possible actions are sampled from the generative model, $\{a_i \sim G_w(s)\}_{i=1}^n$, perturbed, and the action with the maximum $Q$-value is selected, giving the BCQ policy:
\begin{align}
    \pi_{BCQ}(s) = \argmax_{a_i+\xi(s,a_i;\Phi)}Q_{\theta}(s,a_i+\xi(s,a_i;\Phi))
\end{align}
We focus on the scenario where a model of $p(a|s)$ can be obtained through MLE training on data of known action sequences. This prior model provides a more robust estimate of $p(a|s)$ than one learned from the batch data, assuming the size of the batch is small relative to unsupervised data related to the problem (i.e. when the batch comes from human interaction data). We propose an adaptation of BCQ to discrete action spaces (\textit{DBCQ}) which leverages such a strong pre-trained prior model as an improved version of $G_w$. Since BCQ relies on Double Clipped $Q$-learning \cite{fujimoto2018addressing}, here we use dropout-based uncertainty estimates as in Eq. \ref{eq:dropout_mc}. Because the action space is discrete, we do not use a perturbation model to modify actions, but instead define the DBCQ policy as:
\begin{align}
    \label{eq:dbcq}
    \pi_{DBCQ}(s) = \argmax_{a_i \sim p(a|s)}Q_{\theta_{\pi}}(s,a_i)
\end{align}

\subsection{KL Control from pre-trained prior}
Rather than simply sample from the prior, we would like the $Q$-learning algorithm to directly incorporate the prior into the policy. Thus, we use KL-control to penalize divergence between the prior $p(a|s)$, and the $Q$-network policy $\pi_{\theta}$, while still maximizing reward.
Given a trajectory of actions, $\tau=\{a_1, a_2, ... a_{t-1}\}$, let $q(\tau) = \prod_{t=1}^T \pi_{\theta}(a_t, s_t)$ be the policy of our $Q$-learning algorithm at the trajectory level. Similarly, let $p(\tau) = \prod_{t=1}^T p(a_t|s_t)$ be the prior distribution over the trajectory, and $r(\tau)$ be the rewards. We seek to maximize the following KL-regularized objective:
\begin{align}
    L(q) = \mathbb{E}_{q(\tau)}[r(\tau)]/c - D_{KL}[q(\tau)||p(\tau)]
\end{align}

Since $D_{KL}[q||p] = \sum_x q(x)(\log q(x)-\log p(x))$, we can see that this is equivalent to maximizing the following expected value function of the policy $\pi_{\theta}$ at the action level:
\begin{align}
    \vspace{-0.5cm}
    \label{eq:kl_value}
    Q^{\pi}(s_t, a_t) = \mathbb{E}_{\pi}[\sum_{t'=t}^T r(s_{t'}, a_{t'})/c + \log p(a_{t'}|s_{t'}) - \log \pi(a_{t'}|s_{t'})]
    \vspace{-0.5cm}
\end{align}
The two terms we have introduced in Eq. \ref{eq:kl_value} have clear motivations. The $p(a|s)$ term rewards the model for choosing actions that have high probability under the prior, biasing 
the model to state-action pairs that are realistic, and likely to be in the batch. 
The $-\log \pi(a|s)$ term is analogous to entropy regularization. Maintaining diversity in the action space through entropy regularization is important for generative models like dialog systems, which are known to collapse to an uninteresting, small number of repeated samples \cite{li2016dialogue}. Re-stating Eq. \ref{eq:kl_value} as an entropy-regularized $Q$-function, we obtain:
\begin{align}
    Q(s_t, a_t) = \mathbb{E}_{\pi}[\sum_{t'=t}^T r(s_{t'}, a_{t'})/c + \log p(a_{t'}|s_{t'}) + \mathcal{H}(\cdot|s_{t'})]
\end{align}
Motivated by energy-based models of the form $\pi(a_t|s_t) \propto \exp(-\mathcal{E}(s_t,a_t))$, one can derive a soft version of the entropy-regularized $Q$-function that uses a Boltzmann distribution to estimate future reward \cite{haarnoja2017reinforcement}. We refer to it as a $\Psi$-function following previous work \cite{jaques2017sequence}, which derived this function as a generalization of the $\Psi$-learning proposed by \cite{rawlik2012stochastic}. The optimal $\Psi$-function and policy are:
\begin{align}
    \Psi^*(s_t, a_t) &= r(s_{t'}, a_{t'})/c + \log p(a_{t'}|s_{t'}) + \gamma \log \sum_{a'}\exp(\Psi^*(s', a')) \\
    \pi^*_{\Psi}(a_t|s_t) &= \exp(\Psi^*(s_t, a_t))
\end{align}
Because it avoids taking a hard max over noisy estimates, $\Psi$-learning leads to less overestimation of future reward \cite{abdolmaleki2018maximum, haarnoja2017reinforcement}. This leads to more stable TD updates and aids learning. Thus, we argue it will be especially useful in the BRL setting for reducing optimism in the face of uncertainty. 

\subsection{Model averaging}
Finally, we explore the setting where the data in the batch may be generated from a large variety of different models $M$ with different architectures, which each learn a different estimate of $p(a|s;M)$. We use this diversity to create a more robust prior by computing a weighted average of these models based on a normalized score $S(M)$ for each model. The score could be some measure of model quality, or simply the proportion of data in the batch that was generated with that model. Thus, we define $p_{MA}(a|s)$ as the model-averaged prior: $p_{MA}(a|s) = \sum_M S(M) p(a|s;M)$. 

\section{RL for open-domain dialog generation}
\begin{wrapfigure}{r}{0.5\textwidth}
  \vspace{-0.3cm}
  \begin{center}
    \includegraphics[width=0.5\textwidth]{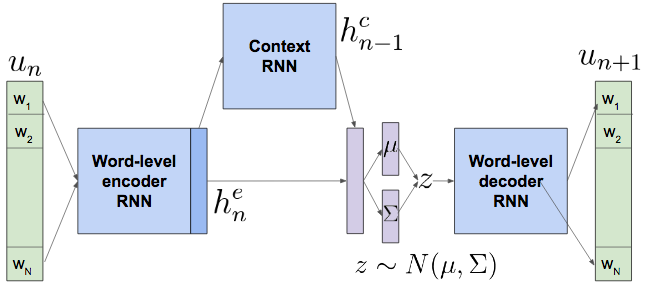}
    \caption{Simplified diagram of the variational hierarchical dialog model.}
    \label{fig:models}
  \end{center}
\vspace{-0.3cm}
\end{wrapfigure} 

In this work, we employ hierarchical seq2seq dialog models \cite{ghandeharioun2019approximating,park2018hierarchical,serban2016building,serban2017hierarchical}, which use three recurrent networks to generate the next utterance in a conversation (see Figure \ref{fig:models}). The \textit{encoder RNN} operates on the tokens of the next input utterance $u_t = [w_1, w_2,... w_n]$, and encodes them into a representation $h^e_t = f^e(u_t)$. This is fed into a \textit{context RNN}, which forms the upper level of the hierarchy -- it is updated only after each utterance, rather than each token. 
The context RNN outputs $h^c_t = f^c(h^e_t)$, which is fed into the \textit{decoder RNN}, which produces the output utterance $u_{t=1}$ one token at a time. Note that while transformer architectures (e.g. \cite{radford2019language}) have emerged as a powerful alternative to seq2seq models, here we choose to focus on hierarchical architectures because it gives us the flexibility to extend this work to use hierarchical control in the future, by learning to optimize rewards at both the utterance and conversation level. Although we trained and tested a variety of different architectures drawing from several works \cite{ghandeharioun2019approximating, park2018hierarchical,serban2016building,serban2017hierarchical}, we converged on the Variational Hierarchical Recurrent Encoder Decoder (VHRED) as the most promising model \cite{serban2017hierarchical}. We also apply knowledge distillation to improve the model's ability to recognize and encode the sentiment and semantics of the conversation, as proposed by \cite{ghandeharioun2019approximating}. 

Applying RL to dialog generation is challenging due to the large state-action space. 
The model attempts to construct a response utterance $u^{\pi}_t = [a_1, a_2,..., a_n]$ by iteratively choosing an action $a_i$ as the next token. The number of tokens in the vocabulary of our pre-trained model is 20,000, making the action space very high-dimensional, potentially compounding the problem of overestimation and making batch learning excessively difficult. However, initializing the $Q$-networks with the weights of the pre-trained language model provides a strong prior over the appropriate word to select.

Here we consider human interaction to represent the `environment'. The response of a human to the bot's utterance is used to compute a reward signal to train the model. The state of the environment $s_t$ constitutes all of the text in the conversation uttered so far, both by the bot and the human. The state has a hierarchical structure, marking its division into utterances, which are further divided into tokens. While the bot is constructing an utterance  $u^{\pi}_t$, 
it is straightforward to obtain a target $Q$-estimate of future reward using the model's estimated $Q$-values over its own next token in the utterance. However, at the last token of the bot's utterance, the estimated future reward must include the human's response $u^h_t$. Therefore, we append the human response into the conversation, $s_{t+1} = [s_{t-1}, u^{\pi}_t, u^h_t]$, feed this into the target $Q$-network, and use the estimated $Q$-values for the first token of the bot's next utterance. All of the code for our models and RL techniques is available in open-source at \url{https://github.com/natashamjaques/neural_chat/tree/master/rl}.

\subsection{Learning from implicit human preferences}
\label{sec:implicit}
We would like to improve a dialog model's ability to engage in natural conversation with a human by learning from the signals implicit in the way that the human responds. Rather than having the human manually label good performance -- which we show in this work does not scale -- the agent should recognize informative cues within the user's responses, like sentiment, and the amount of time they spend chatting. Essentially, we want to create an agent that is intrinsically motivated to produce positive reactions in its human conversation partner.
We design several intrinsic reward functions based on the rich, interactive content of conversation, taking inspiration from the psychology of human conversation: 1) eliciting positive sentiment and transitions from negative to positive sentiment, due to the importance of emotion to creating a sense of understanding  \cite{bodie2015role, weger2010active}; 2) eliciting longer conversations and more words typed, since this is a signal of engagement \cite{sidner2004look,zhou2018design}; 3) eliciting laughter (counting the number of `ha's in the user response), because of its importance in building solidarity \cite{hay2000functions}; 4) high semantic similarity (close distance in sentence embedding space \cite{conneau2017supervised}) between the human input and agent response, because paraphrasing and style matching are important in facilitating good conversation \cite{ireland2011language, weger2010active}; and 5) asking questions, since this is an important active listening skill \cite{bodie2012listening}. The total reward given to the agent is a combination of these, with details (and coefficients) given in the supplementary material. Note that the first 4 types of rewards depend on eliciting positive responses from a human user; we call these the implicit \textit{human reward}. The 5th reward is easily exploitable by the agent itself. 
These rewards represent only an initial foray into designing good metrics of human enjoyment, and further experimentation will be needed to improve them.

\section{Experiments}
To collect interactive human conversation data, we built a CUDA-capable web app that can host neural network dialog models on GPU for fast, real-time inference: \url{https:neural.chat}. The code for the server is available in open-source at \url{https://github.com/asmadotgh/neural_chat_web}. We trained over 40 dialog models with different architectures (e.g. \cite{serban2017hierarchical,serban2016building,park2018hierarchical,ghandeharioun2019approximating}), on different datasets (movie dialogs \cite{cornell_dataset} and Reddit \cite{ghandeharioun2019approximating}). Note that these models varied significantly in terms of the distribution of language they learned. We collected a batch of data containing 14232 pairs of user input and agent response. 
This batch was used to train the RL models described in Section \ref{sec:methods}, which were then re-deployed to the website. We recruited 90 Mechanical Turk workers to provide a total of 718 7-point Likert scale ratings of the bots' quality, fluency, diversity, contingency (relatedness), and empathy, after interacting with each bot for at least 3 turns. Participants also had the option to provide explicit feedback through upvoting or downvoting a particular utterance within the interface. Note that testing these models in the wild with humans represents a more meaningful test of generalization than testing an RL model in the same limited (game) environment in which it was trained, since humans are not restricted in the text they can type to the model, and are the ultimate authority on naturalistic conversation.

\section{Results}
\label{sec:results}

\begin{table}[t]
\vspace{-0.3cm}
\caption{Interactive human evaluation of techniques for off-policy batch RL. KL-control models strongly out-perform other techniques. Ratings are Likert scale, votes and human reward are $z$-scores.}
\label{tab:techniques_human}
\resizebox{\textwidth}{!}{%
\begin{tabular}{|l|lllll|l|l|l|}
\hline
\textbf{Model type} & \textbf{Quality} & \textbf{Fluent} & \textbf{Diverse} & \textbf{Related} & \textbf{Empathy} & \textbf{Total}  & \textbf{Votes} & \textbf{\begin{tabular}[c]{@{}l@{}}Human\\ reward\end{tabular}} \\ \hline
DBCQ                & 1.64  $\pm$.29         & 1.87  $\pm$.34         & \textbf{3.13 $\pm$.58}  & 1.84  $\pm$.34          & 2.09  $\pm$.38          & 10.58 $\pm$1.55          & -.228          & -.050 \\
Batch Q             & 1.87  $\pm$.30          & 2.36 $\pm$.42           & 2.20 $\pm$.41            & 1.91 $\pm$.32           & 2.58 $\pm$.47            & 11.91 $\pm$1.58         & -.163          & -.005 \\
Batch Q + MC        & 1.85 $\pm$.39            & 2.46 $\pm$.44           & 2.46 $\pm$.52            & 1.98 $\pm$.39            & 2.34 $\pm$.49            & 11.07 $\pm$1.82          & -.068          & .005  \\
KL-control Q        & 2.38 $\pm$.39            & 3.24 $\pm$.47           & 3.42 $\pm$.54            & 2.38 $\pm$.45            & 2.56 $\pm$.43            & 13.98 $\pm$1.81          & .016          & .004  \\
KL-control $\Psi$   & 2.33 $\pm$.41            & \textbf{3.73 $\pm$.53}  & 2.82 $\pm$.50            & 2.31 $\pm$.44            & \textbf{3.47 $\pm$.50}   & \textbf{14.67 $\pm$1.82} & \textbf{.128}  & \textbf{.061} \\
KL-control MA $\Psi$    & \textbf{2.60 $\pm$.43}   & 3.47 $\pm$.42           & 3.00$\pm$.49            & \textbf{2.49 $\pm$.44}   & 2.89 $\pm$.51            & 14.44 $\pm$1.96          & .127           & .042  \\ \hline
\end{tabular}
}%
\vspace{-0.1cm}
\end{table}

To compare models, we not only look at human users' ratings and votes, but also consider the automatic signals detectable from the text itself. This implicit \textit{human reward} metric aggregates the measures listed in items 1-4 in Section \ref{sec:implicit}, and measures the ability to elicit positive responses from the human. Table \ref{tab:techniques_human} shows the results of the evaluation. Each of the enhancements proposed (MC estimation of target $Q$-values, $\Psi$-learning, and MA) leads to performance gains in terms of human reward, manual votes, or ratings. 
However, the most notable difference in performance comes from KL-control. The KL-control models show substantial gains over the baseline\footnote{We also compare the RL models to the prior, and see performance improvements in terms of the elicited human reward, but not in terms of the quality ratings. We believe this is because the rewards proposed here do not fully cover what it means to have a high quality conversation (as we will elucidate later), and hope that other researchers be able to use the techniques we propose to learn from improved rewards.}
models across both ratings and human reward. We perform a one-way analysis of variance (ANOVA) comparing the KL-control models to the Batch Q baselines and DBCQ on the total human rating score, and find that the KL-control models are significantly better, $F(x) = 4.781, p <.05$. This validates the hypothesis that KL-control with a strong, pre-trained prior can be used to improve batch RL. 


\begin{table}[t]
\vspace{-0.1cm}
\caption{Purely reward-maximizing methods like \textit{Batch Q} trivially exploit the reward function by asking a question every turn, and using the maximum number of tokens in every sentence. In contrast, KL-control methods output plausible language by staying close to the prior, but shift to using polite, cheerful language to maximize implicit human rewards.}
\vspace{0.1cm}
\label{tab:samples}
\resizebox{\textwidth}{!}{%
\begin{tabular}{l|l}
\begin{tabular}[c]{@{}l@{}}
    \textbf{{[}User{]}}: hey, what's up? \\ 
    \textbf{{[}Batch Q{]}}: where did you get??, that sounds so exciting. how do you work for you... \\
         you want to talk to me?... you want\\
    \textbf{{[}User{]}}: yes, i want to talk to you! \\ 
    \textbf{{[}Batch Q{]}}: where did you say to me?... you want to talk to you about you...\\
     you are so good for you... you are so good...\\
    \textbf{{[}User{]}}: thanks, i appreciate that. \\
    \textbf{{[}Batch Q{]}}: where did you get??, that sounds so exciting. how do you work for you...\\
     you want to see you...
\end{tabular} & 
\begin{tabular}[c]{@{}l@{}}
    \textbf{{[}User{]}}: hey, what's up? \\ 
    \textbf{{[}KL-control{]}}: i'm not sure, but i'm glad to see you, and \\
    i'm so happy that you're not alone\\ 
    \textbf{{[}User{]}}: oh thanks, that's very nice of you! \\ 
    \textbf{{[}KL-control{]}}: i'm sorry, but i'm glad to see what you think. \\ 
    \textbf{{[}User{]}}: don't be sorry, you did a good job. \\ 
    \textbf{{[}KL-control{]}}: that's really nice to me. 
\end{tabular} 
\end{tabular}
}
\vspace{-0.1cm}
\end{table}

\begin{wrapfigure}{r}{0.5\textwidth} 
    \vspace{-0.3cm}
  \begin{center}
   \includegraphics[width=.5\textwidth]{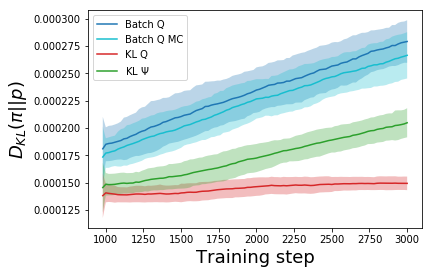}
  \caption{KL-divergence of the policy from the prior is lower with KL-control throughout training. Bands show standard deviation.}
  \label{fig:kl}
  \end{center}
  \vspace{-0.3cm}
\end{wrapfigure} 

Without KL-regularization, the baseline RL models diverge quickly and continuously from the prior, losing information about realistic sequences -- as shown in Figure \ref{fig:kl}. This figure also helps explain the poor performance of DBCQ in Table \ref{tab:techniques_human}. The underlying $Q$-network in DBCQ does not directly integrate the prior. As $Q$-learning causes the model to diverge from the prior, the $Q$-estimates of language generated according to the prior become unrealistic, and Eq. \ref{eq:dbcq} selects unrealistic actions. This results in highly `diverse' (random) generated utterances. Note that since we operate in discrete action space, we could not include the perturbation model originally proposed by \cite{fujimoto2018off}, which may be critical to achieving good performance with BCQ.

\begin{figure}
  \centering
  \includegraphics[width=.6\textwidth]{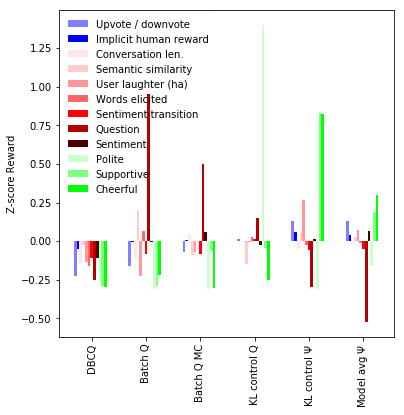}
  \caption{Z-scored reward. Red metrics were used in training rewards, green are post-hoc. Traditional RL methods like Batch Q exploit simple action-based rewards, like asking questions. In contrast, KL-control methods shift their distribution towards polite, supportive, and cheerful conversation, allowing them to elicit higher \textit{human reward} (blue).}
  \label{fig:technique_rewards}
  \vspace{-0.1cm}
\end{figure} 

The pre-trained prior may be especially important in a generative domain like dialog, where the true reward function is unknown, and so purely maximizing reward may actually lead to poorer quality conversations. Table \ref{tab:samples} shows examples of conversations with a Batch $Q$ and KL-control model. Because the Batch $Q$ model has no incentive to stay close to realistic language, it learns to exploit the reward by asking a question and outputting the maximum number of tokens (30) every utterance. These sentences contain implausible phrases that do not represent realistic language (e.g. ``\textit{where did you say to me?}"). In contrast, the KL-control model uses realistic language, but shifts its distribution towards cheerful and polite speech, presumably because this is what led to positive human responses in the batch data. Rather than simply cherry-picking results, we invite the reader to check for themselves; all of the models tested in this study are available at: \url{https://neural.chat/rl}.

In fact, we noticed that all models trained with the implicit human rewards described in Section \ref{sec:implicit} learned to use more cheerful and supportive language. Therefore, we create post-hoc metrics to measure this effect (see the supplementary material for details). Figure \ref{fig:technique_rewards} shows how these metrics, as well as the implicit rewards, differ across models. Without KL-control, baseline methods like Batch Q exploit simple rewards like asking questions at the expense of realistic language, explaining their poor quality ratings. In contrast, KL-control models learn to rely more on realistic but polite, supportive, and cheerful dialog to elicit higher total \textit{human reward}.

\begin{figure*}[t]
\vspace{-0.1cm}
  \centering
  \begin{subfigure}[t]{0.25\textwidth}
  \includegraphics[width=\textwidth]{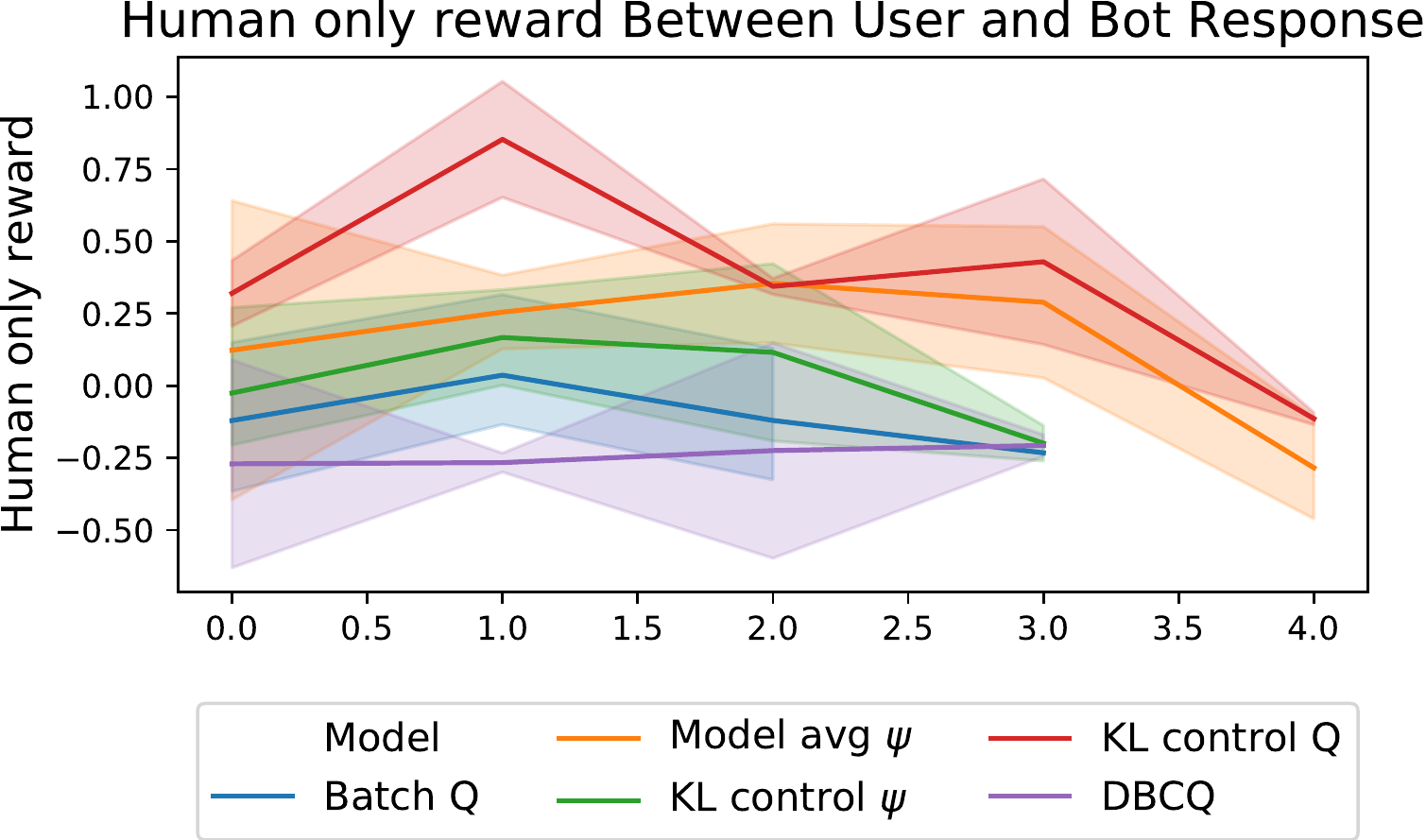}
  \caption{}
\end{subfigure}
\begin{subfigure}[t]{0.24\textwidth}
  \includegraphics[width=\textwidth]{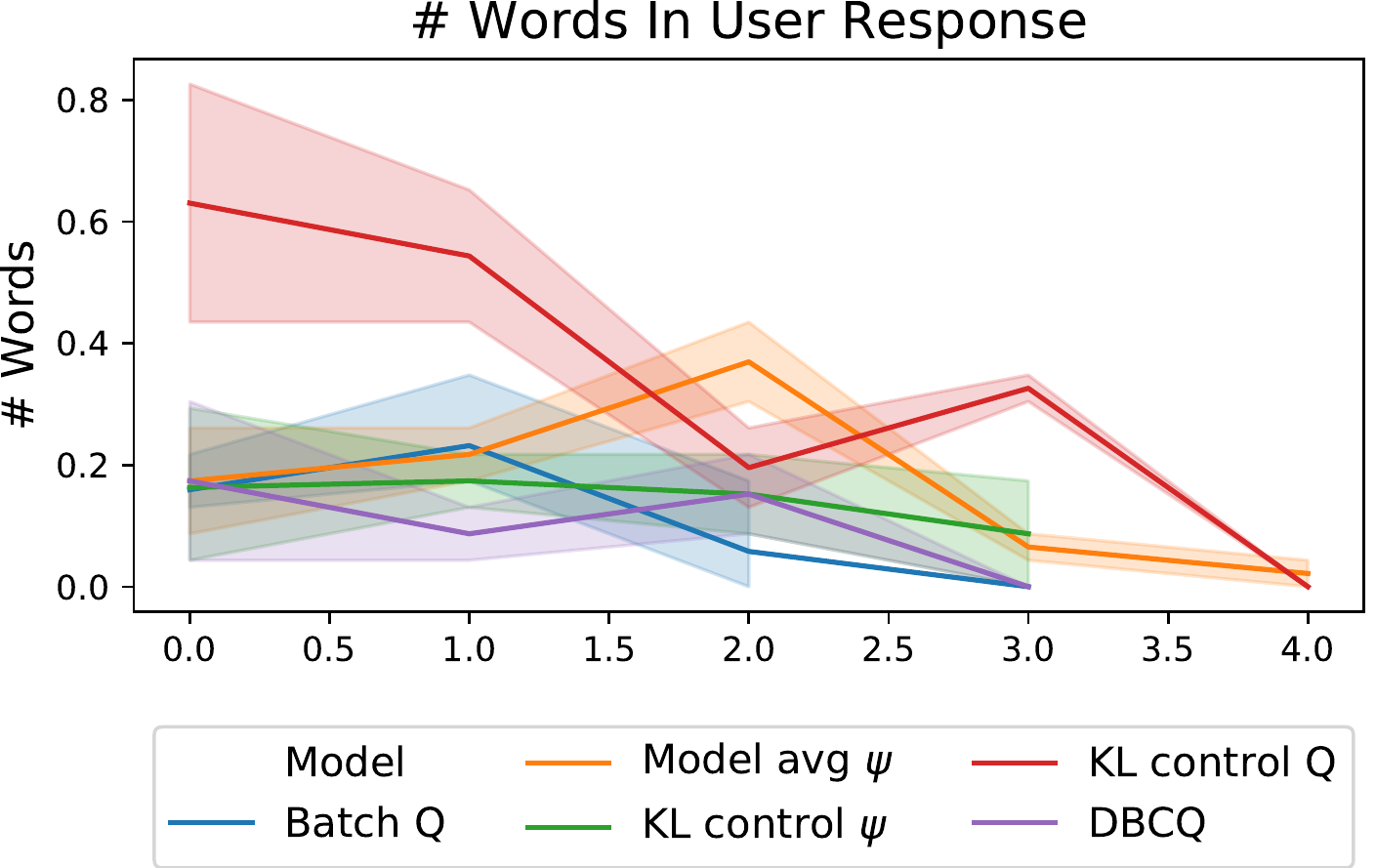}
  \caption{}
\end{subfigure}
\begin{subfigure}[t]{0.24\textwidth}
  \includegraphics[width=\textwidth]{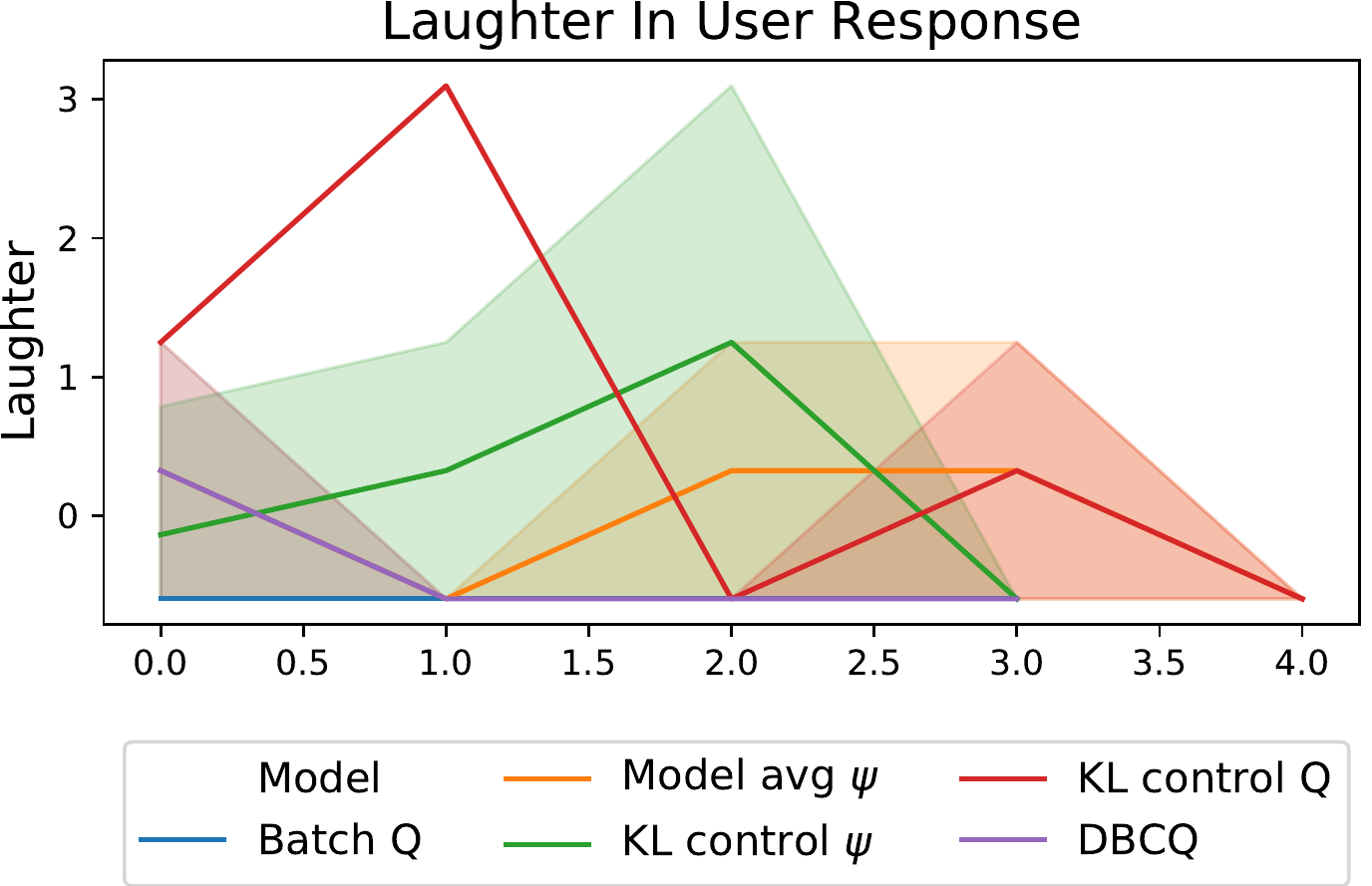}
  \caption{}
\end{subfigure}
\begin{subfigure}[t]{0.25\textwidth}
  \includegraphics[width=\textwidth]{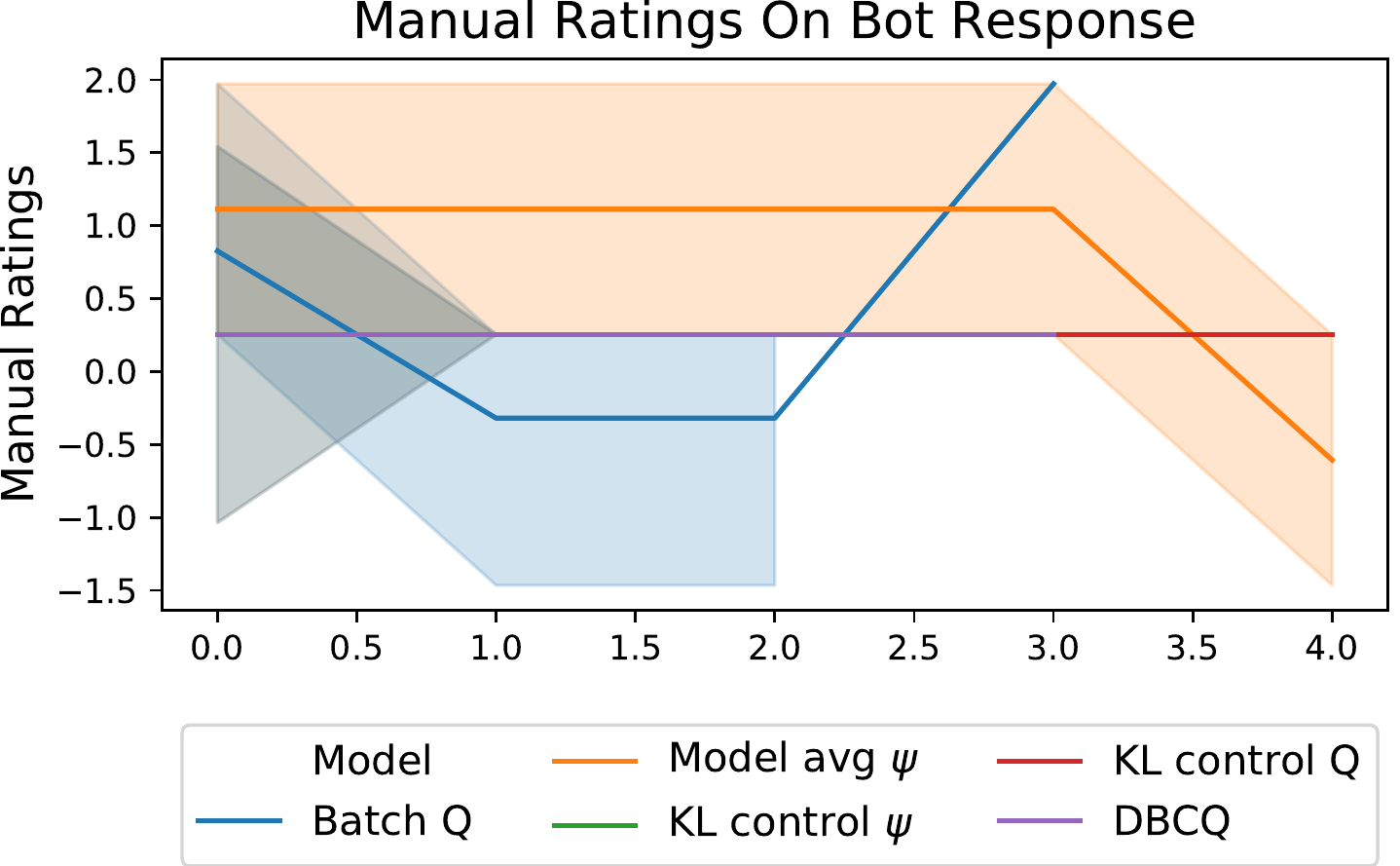}
  \caption{}
\end{subfigure}
\caption{Comparison of top 10 conversation trajectories observed across deployed models, 90\% CI of the rewards: (a) Implicit human feedback; (b) Words elicited; (c) Laughter; (d) Manual votes.}
  \label{fig:trajectories}
\vspace{-0.2cm}
\end{figure*}

To understand the effect of the implicit rewards, Figure \ref{fig:trajectories} shows the reward trajectory over the ten best conversations obtained with models trained with different techniques. While we see that KL-control models are able to elicit significantly higher reward than baselines, we note that KL-control $Q$ performs best overall and in terms of words elicited, even though it had lower quality ratings in Table \ref{tab:techniques_human}. 
This suggests that maximizing these rewards is not a perfect proxy for human judgments of quality. Note also that eliciting laughter is an extremely rare event, and only the KL-control models are able to do so. Finally, Figure \ref{fig:trajectories} (d) shows that manual votes occur even more rarely, suggesting that explicit feedback from humans is a cumbersome and sparse reward signal. 

Table \ref{tab:rewards_human} presents the results of models trained with only a single reward function, ordered from lowest to highest quality. Notably, extracting multiple different reward functions post-hoc from a batch of data and training on these independently is only possible with an effective BRL model. Here all models are trained with KL-control, $\Psi$-learning, and MC targets. Investigating which rewards presented in Section \ref{sec:implicit} are most critical to achieving high-quality conversations with humans, we note that maximizing positive and minimizing negative sentiment in the user turns out to lead to the highest quality bot. This underscores the importance of affective signals as cues for good conversation. Bots trained on the manual upvotes and downvotes provided by users on the utterance level fail to achieve similarly high performance. Even though users were instructed to make use of the vote feature, the task is burdensome, and users did not vote frequently enough to provide a good training signal. This validates the hypothesis that \textit{implicit} signals of human enjoyment (such as sentiment) are a more scalable way to learn from human preferences.

\begin{table}[t]
\centering
\caption{Interactive human evaluation of different reward functions (models trained with KL-control)}
\label{tab:rewards_human}
\resizebox{\textwidth}{!}{%
\begin{tabular}{|l|lllll|l|l|l|}
\hline
\textbf{\begin{tabular}[c]{@{}l@{}}Reward\\ function\end{tabular}} & \textbf{Quality} & \textbf{Fluent} & \textbf{Diverse} & \textbf{Related} & \textbf{Empathy} & \textbf{Total}  & \textbf{Votes} & \textbf{\begin{tabular}[c]{@{}l@{}}Human\\ reward\end{tabular}}  \\ \hline
Conv. len.      & 2.20 $\pm$.40            & 3.61 $\pm$.53          & 3.02 $\pm$.52           & 2.25 $\pm$.46   & 2.48 $\pm$.45           & 13.57 $\pm$1.84         & -.035          & -.003  \\
Semantic sim.   & 1.93 $\pm$.34            & 3.50 $\pm$.45           & 2.37 $\pm$.45            & 2.11 $\pm$.45            & 2.52 $\pm$.48            & 12.43 $\pm$1.75          & -.020          & .012  \\
User laughter   & 1.96 $\pm$.38            & 3.56 $\pm$.48           & 2.33 $\pm$.51            & 1.93 $\pm$.42            & 3.20 $\pm$.55            & 12.98 $\pm$1.60          & -.149          & -.003 \\
Words elicited  & 2.11 $\pm$.32            & 3.96 $\pm$.44           & 3.04 $\pm$.45            & 2.04 $\pm$.35            & 2.55 $\pm$.46            & 13.70 $\pm$1.44          & .059           & .024  \\
Manual votes    & 2.14 $\pm$.38            & 3.47 $\pm$.45           & 2.91 $\pm$.47            & 2.07 $\pm$.39            & 2.42 $\pm$.46            & 13.00 $\pm$1.65          & -.030          & .010 \\
Sent. trans.    & 2.02 $\pm$.31            & 3.71 $\pm$.49           & 2.98 $\pm$.50            & 2.04 $\pm$.42            & 2.84 $\pm$.48            & 13.60 $\pm$1.63          & .031           & .014  \\
Question        & 2.29 $\pm$.37            & \textbf{4.31 $\pm$.50}  & \textbf{3.31 $\pm$.52}   & 2.20 $\pm$.40            & 2.60 $\pm$.41            & 14.71 $\pm$1.63          & .057           & .012  \\
Sentiment       & \textbf{2.47 $\pm$.32}   & 4.05 $\pm$.45           & 3.23 $\pm$.46            & \textbf{2.42 $\pm$.39}            & \textbf{3.23 $\pm$.55}   & \textbf{15.40 $\pm$1.49} & \textbf{.085}  & \textbf{.045}  \\ \hline
\end{tabular}%
}
\vspace{-0.5cm}
\end{table}

\section{Conclusion}
This paper presents a series of techniques which improve performance when learning off-policy without the possibility to explore -- i.e. batch RL (BRL). Most significantly, we are the first to propose using KL-control from a strong prior model pre-trained on data as a way to avoid overestimation and instability in BRL. Our results demonstrate that KL-control is critical to achieving good performance in this setting. In a generative domain such as dialog, the true reward function is not known, and trivially exploiting the rewards can actually lead to worse performance. Thus, KL-control may be particularly necessary to ensure samples remain realistic and close to the data distribution. We propose several reward functions that could allow an open-domain dialog generation model to learn from rich cues implicit in human interaction, where learning from expressed sentiment was most promising. While these rewards are far from perfect or complete, we see that maximizing implicit rewards leads to better performance than relying on explicit feedback. We hope that the techniques presented here will allow other researchers to leverage BRL for learning from human interaction data, and spur the development of even better rewards for capturing human preferences.

 \subsubsection*{Acknowledgments}
We would like to thank Scott Fujimoto for insightful email correspondence on this topic, approval of the DBCQ algorithm, and suggestion to apply model averaging. We also thank Max Kleiman-Weiner, Ardavan Saeedi, Sebastian Zepf, Sara Taylor, Oliver Saunders Wilder, Kyle Kastner, and Kristy Johnson for their helpful discussions about this project, and many others for helping test-drive our bots. %

We thank the MIT Quest for Intelligence, and MIT Stephen A. Schwarzman College of Computing, and the Machine Learning Across Disciplines Challenge for providing computing resources, and MIT Media Lab Consortium for the support of this research.

\small
\bibliographystyle{plain}
\bibliography{dialog}

\section{Appendix}

\subsection{Details about implicit metrics}

\subsubsection{Sentiment-based}
\begin{figure*}[h]
  \centering
  \begin{subfigure}[t]{0.5\textwidth}
  \includegraphics[width=\textwidth]{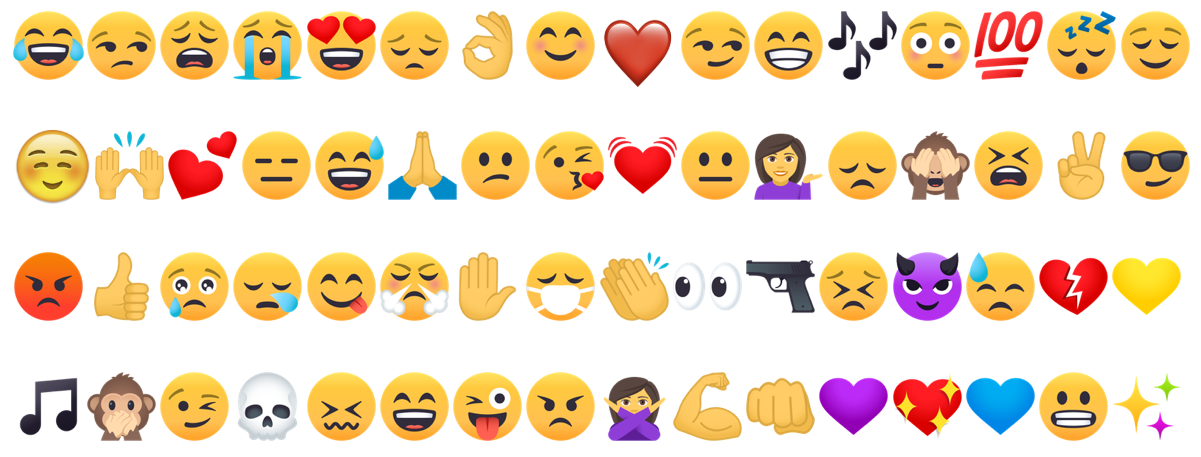}
  \caption{}
  \end{subfigure}
  \hspace{1cm}
  \begin{subfigure}[t]{0.35\textwidth}
  \includegraphics[width=\textwidth]{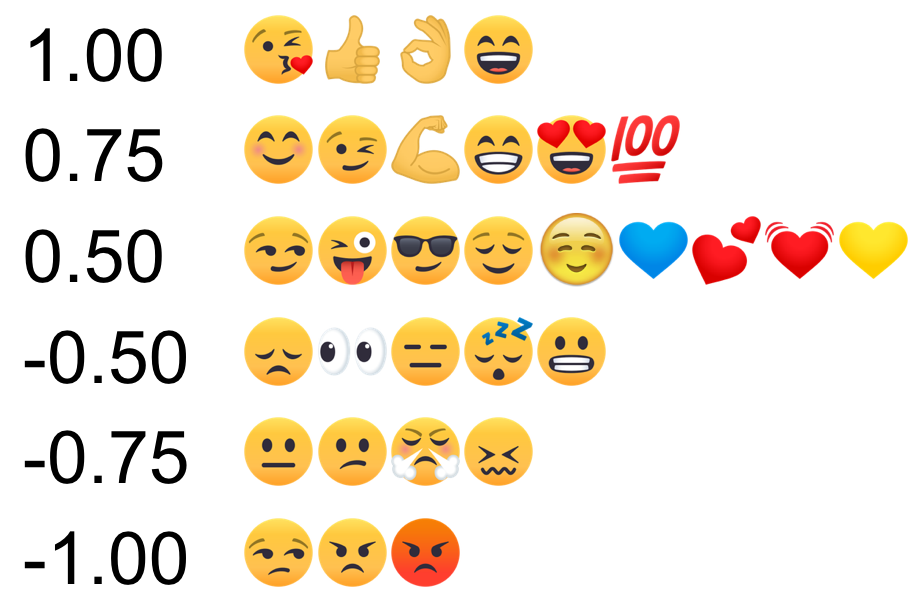}
  \caption{}
  \end{subfigure}
  \caption{(a) 64-most frequent emojis as predicted by \cite{felbo2017using} used for calculating emotion embeddings. (b) Assigned weights used in producing the sentiment reward from the predicted emoji values.}
  \label{fig:emojis}
\end{figure*}

To compute sentiment on short texts like conversation utterances, we leverage a state-of-the-art sentiment-detection model, which was trained on a massive amount of Twitter data to predict the emojis in tweets \cite{felbo2017using}. Transfer learning from this model to other tasks showed that it was able to significantly outperform a series of sentiment, irony, and sarcasm benchmarks. This DeepMoji model outputs a probability distribution over 64 most-frequently used emojis as shown in Figure \ref{fig:emojis}. After observing the performance of the model in detecting users' emotions in the domain of online chat, we define a set of weights over the emojis and calculate the weighted sum over an emotion embedding vector to derive a \textit{sentiment} reward which is higher for positive sentiment and lower for negative sentiment. These weights are shown in Figure \ref{fig:emojis} (b). We also compute a sentiment-transition reward using the same score based on whether the peak positive sentiment occurred later in the conversation than the peak negative sentiment, reasoning that sentiment should improve over the course of the conversation. 

\subsubsection{Engagement-based}
Based on prior work \cite{zhou2018design}, we use the number of turns in the conversation as an indicator of the quality of the bot's performance. To distribute this reward over every utterance in the conversation, we take the total conversation length $N$, and compute the discounted reward for utterance $n < N$ as $\gamma^{N-n}N$. We also reward each utterance with the number of words in the user's response, which we refer to as the \textit{words elicited}.

\subsubsection{Laughter}
Laughter has been shown to be very important to human affiliation \cite{provine1996laughter} and solidarity \cite{hay2000functions}. Therefore, we detect the number of occurrences of the string `ha' in the user's response, and use this as a reward. Interestingly, we find that bots trained to maximize user laughter learn to be extremely supportive and cheerful compared to other bots (for definitions of supportive and cheerful, see Section \ref{sec:posthoc}).

\subsubsection{Semantic similarity}
Language style matching has been shown to be a strong predictor of relationship initiation and stability \cite{ireland2011language}. While it would be ideal if our chatbots could intelligently adapt their conversation style to a new user, in reality most baseline dialog models struggle to maintain topic coherence, even over a few utterances (for an analysis of this effect, see \cite{ghandeharioun2019approximating}). Therefore we reward \textit{semantic similarity} between the user's input and the bot's response, to encourage the bot to stay on topic and produce reasonable answers. This score is computing by leveraging a state-of-the-art sentence embedding model \cite{conneau2017supervised}, and penalizing distance in embedding space.

\subsubsection{Questions}
Asking questions is an important listening skill, and is linked to conversation management, attentiveness, and responsiveness \cite{bodie2012listening}. Therefore, we give the bot a reward of 0.5 if the utterance contains a question word (\textit{how, what, where, why, when, who}), and an additional 0.5 if it contains a question mark.

\subsubsection{Total reward equation}
The total reward used to train the bots is a combination of the above rewards, in the following proportions: 

\texttt{0.15682657*question + 0.13837638*semantic\_coherence +  0.15313653*laughter + 0.14206642*sentiment\_transition + 0.14206642*sentiment + 0.14760148*words\_elicited + 0.1199262*conversation\_length}.

\subsubsection{Post-hoc metrics}
\label{sec:posthoc}
After training the bots on these rewards, we noticed a shift in the distribution of their language towards more polite, cheerful, and supportive speech. Therefore, we designed post-hoc metrics to measure these qualities, which are based on counting whether a subset of phrases is present in an utterance.

\textbf{Politeness phrases:} \textit{if I may; may I; please; thanks; no worries; if you don't mind; have a great day; I'm sorry}.

\textbf{Supportive phrases:} \textit{you're right; you are right; you're not alone; you are not alone; congrats; that's a good idea; that is a good idea; you'll be fine; you will be fine; you'll be okay; you will be okay; it will get better; sorry you're going through; sorry you are going through; if it makes you feel better; if it makes you feel any better; keep your head up; keep it up; I'm in a similar situation; I am in a similar situation; you'll get it; you will get it; happy for you; I'm in the same boat; I am in the same boat; if you feel like you need to vent}.             

\textbf{Cheerful phrases:} \textit{nice to hear; happy; excited; really nice; glad; the best; great; good time; looking forward; beautiful}.

\subsection{Training details and hyperparameters}
RL models were trained for between 800 and 1000 batches of data, where the batch size was fixed at 32. Early stopping was used to determine the number of training iterations of the best checkpoint. All other hyperparameters were shared between RL models, and were as follows: discount $\gamma=0.5$, weight placed on RL reward vs. KL-divergence term $c=2$, number of Monte Carlo samples of the Target $Q$-network $M=5$, target network update rate $\alpha=.005$, learning rate $r=.0001$. We used a smooth $L1$ loss function to approximate the $Q$-values, and clipped gradients at a value of $1.0$. 

The underlying parameters of the VHRED model were as follows: Context RNN hidden size $=1000$, decoder hidden size $=1250$, encoder hidden size $=1250$, $z$ embedding size $=600$, gradient clip $=1.0$, dropout $d=0.2$. The maximum conversation length was fixed at 5 utterances (context from more than 5 utterances ago was discarded), and the maximum sentence length was 30 tokens. 

We also added layers to the Context RNN and regularized it to be able to predict the semantic content of the input utterance using a form of knowledge distillation \cite{hinton2015distilling} from a state-of-the-art sentence-embedding model \cite{conneau2017supervised}. There were 2 additional feedforward semantic prediction prediction layers of size 128, which used ReLu activation. 

\subsection{Additional results}
\begin{figure}
  \centering
  \includegraphics[width=\textwidth]{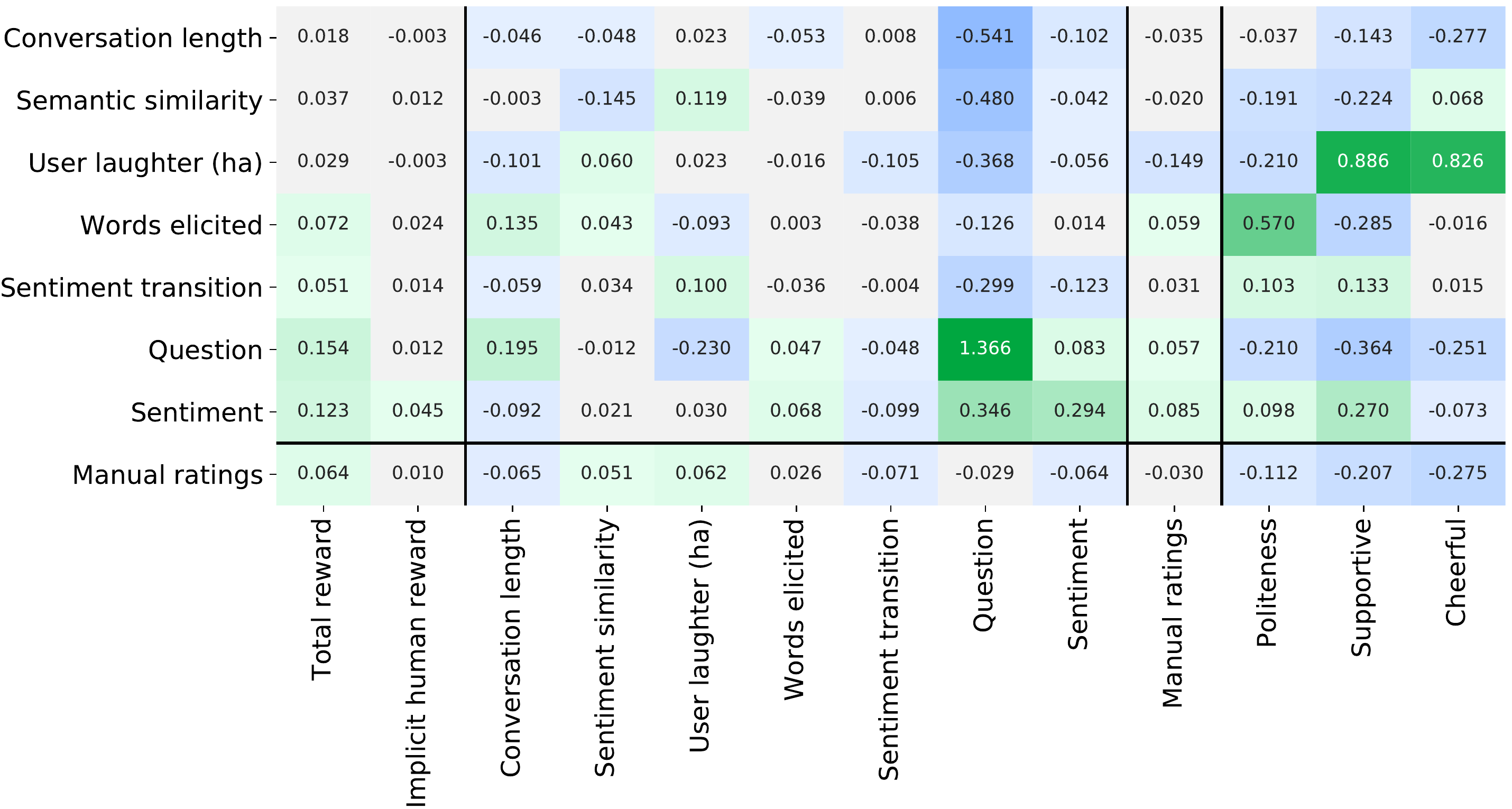}
  \caption{Normalized reward scores obtained by models trained with respect to different rewards. We see that the bot trained to ask questions is easily able to exploit this reward, and similarly the bot trained to elicit positive sentiment does so successfully. For the rest of the bots, the relationship is less clear. For example, the bot trained to elicit laughter becomes the most supportive and cheerful, while the bot trained to elicit more words is very polite.}
  \label{fig:rewards_heatmap}
\end{figure}

Figure \ref{fig:rewards_heatmap} shows the normalized reward scores obtained bots trained with respect to different rewards. While some bots (such as those trained to ask questions or elicit positive sentiment) effectively generalize to new users, we see that others (e.g. words elicited) are not actually able to best elicit those responses in the wild. We hypothesize this is because the relatively small size of batch date we were able to collect ($\approx 14,000$ utterances) does not give these bots enough information about how to elicit long responses from users. 

\subsection{Interactive bot platform details}
\label{sec:appendix-interactive}
\begin{figure}[h]
  \centering
  \includegraphics[width=\textwidth]{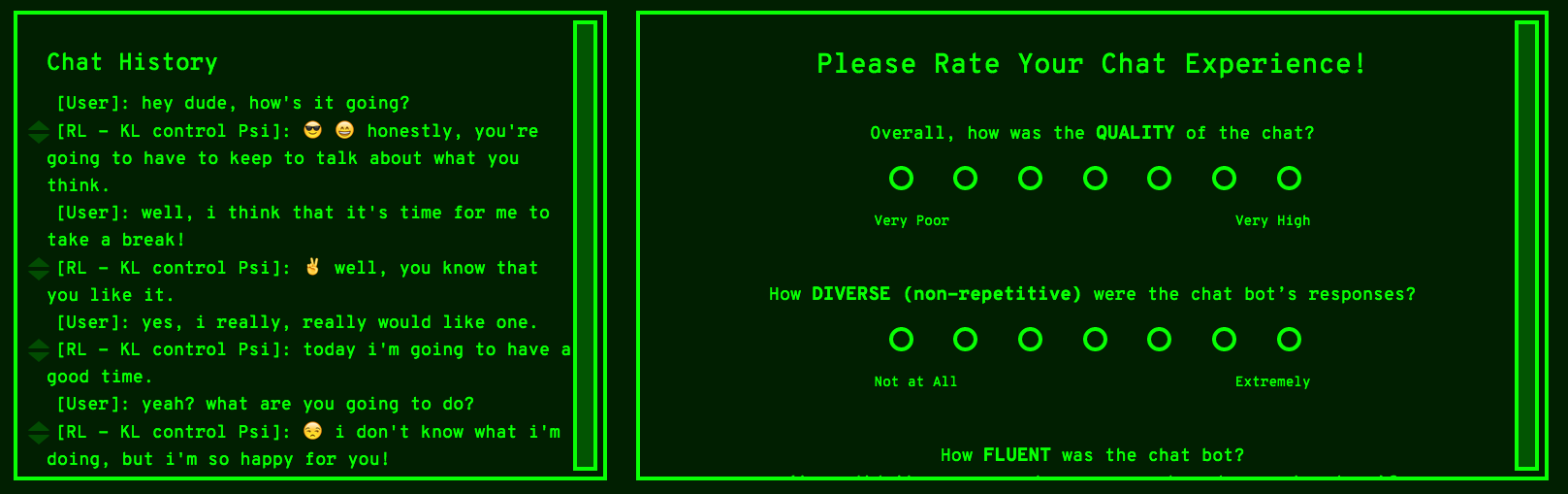}
  \caption{Interactive evaluation ratings page available at \url{https://neural.chat}.}
  \label{fig:interactive-rating}
\end{figure}

To collect data from humans interacting with our bots, we built \url{https://neural.chat}, a platform for hosting deep neural network dialog models online on GPU for fast, real-time inference. Figure \ref{fig:interactive-rating}) shows an example of the interface, in which users are able to rate the bots after talking to them for at least three turns. 

\begin{figure}[h]
  \centering
  \includegraphics[width=.8\textwidth]{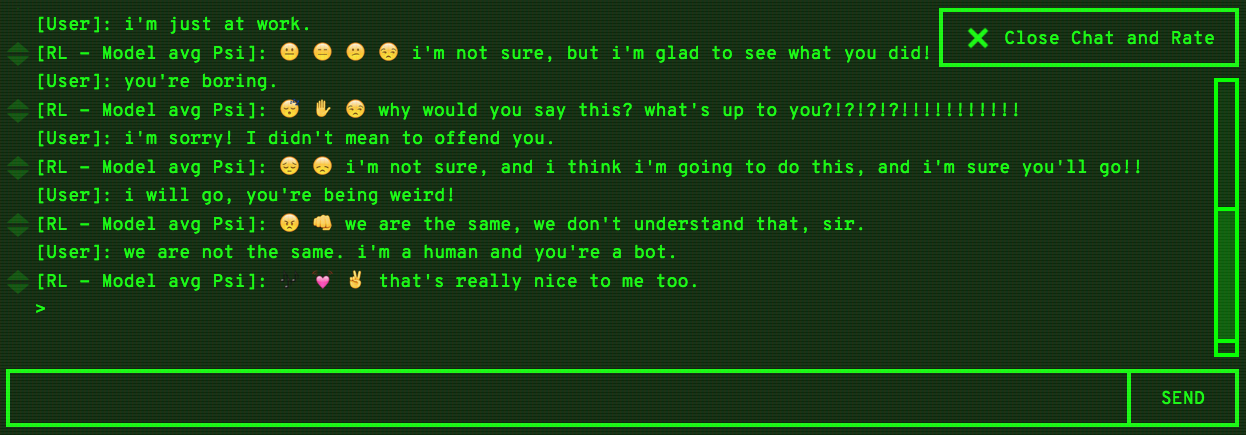}
  \caption{Interactive evaluation chat interface.}
  \label{fig:interactive-chat}
\end{figure}

Figure \ref{fig:interactive-chat} is an example conversation within the platform that interactive evaluation participants see. Annotators can optionally click the up and down arrows beside each chatbot response to give feedback on the specific utterance. Once 3 or more turns of the conversation has taken place, participants may click "Close Chat and Rate" to get to the rating screen. 

\subsubsection{Website server setup and configuration}
\label{sec:appendix-website}

The server was hosted on a Google Cloud Platform virtual instance with 64GB of RAM and a NVIDIA Tesla P100 graphics card. The backend was a Django program being served by NGINX and uWSGI. For simplicity, we opted to have the Django process import the chatbots into the same Python process as Django, rather than have the two connect to each other via other means such as sockets. This configuration decreased development time and increased reliability, but it would need to be revisited if the server needed to scale several orders of magnitude past what was required for this study. The current configuration was still able to support hundreds of simultaneous users and host more than 30 bots concurrently.

The chatbots were kept in a separate project from the Django project and maintained separately from the server code. Each chatbot extended an abstract class that defined key methods for the Django program to use, and was registered to a globally accessible dictionary via a decorator. The Django project was provided the path to the Chatbots project in its PYTHONPATH, so it could import the dictionary in which all the chatbot objects had been registered and use that to dynamically determine which chatbots were available and to access them in its views.

It is important to note that the chatbots used PyCUDA, and PyCUDA does not work in a multiprocessing environment. Because of this, uWSGI needed to be configured to only have one python process and to disable any attempt at multiprocessing. Furthermore, the chatbots required substantial startup times, so all chatbots are kept in memory at all times in the Django process. In order to keep all the chatbots in memory concurrently, we needed a very high amount of RAM on our server and opted for a 64GB virtual instance, and a GPU with 16GB RAM. This combination of CUDA to run the chatbots on the GPU with a high amount of RAM to keep all bots in memory at the same time resulted in incredibly fast server response times, with effectively no increase in response time when using the bots in requests compared to requests that did not.

For further information and instructions on server configuration, please read the server documentation available at \url{https://github.com/asmadotgh/neural_chat_web}. We hope that this platform will allow others to host their own bots and evaluate them in an interactive setting.

\end{document}